\documentclass{article}

\usepackage{microtype}
\usepackage{graphicx}
\usepackage{subfigure}
\usepackage{booktabs} % for professional tables

 \usepackage{hyperref}

\usepackage[accepted]{icml2024}

\usepackage{amsmath}
\usepackage{amssymb}
\usepackage{mathtools}
\usepackage{amsthm}

\usepackage[capitalize,noabbrev]{cleveref}

\theoremstyle{plain}

\theoremstyle{definition}

\theoremstyle{remark}

\usepackage[textsize=tiny]{todonotes}

\icmltitlerunning{Relational Composition}

\begin{document}

\twocolumn[
\icmltitle{Relational Composition in Neural Networks: \\
           A Survey and Call to Action}



\icmlsetsymbol{equal}{*}

\begin{icmlauthorlist}
\icmlauthor{Martin Wattenberg}{hg}
\icmlauthor{Fernanda B. Viégas}{hg}

\end{icmlauthorlist}

\icmlaffiliation{hg}{Harvard University and Google. Work done at Harvard}

\icmlcorrespondingauthor{Martin Wattenberg}{wattenberg@seas.harvard.edu}
\icmlcorrespondingauthor{Fernanda Viégas}{fernanda@seas.harvard.edu}

\icmlkeywords{Machine Learning, ICML}

\vskip 0.3in
]



\printAffiliationsAndNotice{}  % leave blank if no need to mention equal contribution

\begin{abstract}
Many neural nets appear to represent data as linear combinations of ``feature vectors.'' Algorithms for discovering these vectors have seen impressive recent success. 
However, we argue that this success is incomplete without an understanding of \textbf{relational composition}: how (or whether) neural nets combine feature vectors to represent more complicated relationships. To facilitate research in this area, this paper offers a guided tour of various relational mechanisms that have been proposed, along with preliminary analysis of how such mechanisms might affect the search for interpretable features. We end with a series of promising areas for empirical research, which may help determine how neural networks represent structured data.

\end{abstract}

\section{Introduction}

The \textit{linear representation hypothesis} asserts that neural nets encode information as a sum of ``feature vectors'' \cite{mikolov2013linguistic, arora2016latent, olah2020zoom}. That is, a layer of a network represents a set of features\footnote[2]{See Appendix for a definition of the word ``feature.''} as a weighted sum of associated vectors. 
An increasing body of evidence supports this idea, and one recent development is a set of algorithms for automatically finding feature vectors at scale (e.g., \cite{bricken2023towards, cunningham2023sparse, rajamanoharan2024improving, templeton2024scaling} among others). These new techniques raise the possibility that we can understand and even control neural networks using a catalog of interpretable feature vectors.

The very simplicity of the linear representation hypothesis, however, suggests that there may be more to the story. After all, most computer programs contain data structures far more intricate than mere sets or bags, reflecting the complexity of the world. And programs that do treat memory as just a set of bits or bytes, such as assembly language, are notoriously hard to read and interpret. 
It's natural to ask, could there be structure in neural representations beyond this ``bag of features'' model? Indeed, this basic question has attracted notice since the founding of the field \cite{rosenblatt1961principles, fodor1988connectionism}. A wide array of mechanisms have been proposed for relating and composing vector representations---what we refer to here as \textbf{relational composition}. 

This paper includes a summary of some of these mechanisms, with a specific focus on ideas that seem relevant to contemporary interpretability work\footnote[3]{See \cite{schlegel2022comparison} for a more comprehensive historical review of vector compositionality, and \cite{feldman2013neural} for a general survey of the ``binding problem'' in biological neural nets.}. A central goal is to provide a gateway to a set of work that is spread over multiple decades and disciplines. We also analyze how these mechanisms might interact in subtle ways with feature identification algorithms, potentially causing problems for techniques that ``steer'' behavior by intervening during inference. 
In particular, we describe two potential problems that might arise in applying non-compositional methods to compositional representations:
\begin{itemize}

  \item \textbf{Feature multiplicity}. This is a kind of false positive, where combinatorial mechanisms might create multiple distinct ``echo'' feature vectors that correspond to the same concept. 
  Furthermore, despite representing the same concept, different echo vectors potentially could have different effects on the system's output---leading to what we term a predict/control discrepancy. As a result, using feature vectors to understand and control behavior may be challenging.
  \item \textbf{Dark matter}. This possibility represents a false negative. Some combinatorial mechanisms might systematically hide important features or combinations of features. Moreover, by looking for semantically meaningful vectors, we might miss important relational data. For example, binding mechanisms that involve so-called ``ID'' vectors \cite{feng2023language} may encode information not by value, but by reference, leading to ``features'' with no interpretation in terms of system inputs, processing, or outputs.
\end{itemize}

We're certainly not the first to point out the importance of compositionality in this context. Similar issues are raised in \cite{bricken2023towards} and \cite{olah2024distributed}, for example.
Our argument here is that the rich set of existing ideas for compositional representations may hold important clues to understanding recent advances in mechanistic interpretability. Even if they don't capture the full workings of neural nets, these mechanisms might be useful human-interpretable approximations. Examining the literature in this light may help us understand the limitations and results of a linear analysis of neural net representations. Given the potential payoff for new results, our aim in this paper is to accelerate research in this area.

\section{Definitions, notation, and background}

For simplicity, we'll work in an idealized setting, meant to be the bare minimum to explore ideas about composing representations. The basic unit of representation in this setting is a vector $x \in \mathbb{R}^n$. We'll assume that this vector encodes a set of ``features'' using vector addition:
$$x = \sum_{i = 1}^m a_i v_i$$
where each $v_i$ is a unit vector representing an individual feature, and the coefficient $a_i$ represents the weight---or sometimes just presence or absence---of that feature in the representation. (See appendix for  more on the term ``feature.'') Note that we may have $m > n$, that is, more features than dimensions; this is called ``superposition'' in the literature (\cite{elhage2022toy} discusses some of the history of this idea.) In this setting, one typically assumes that features are ``sparse'', meaning most values of $a_i$ are zero, and that the $v_i$ are nearly mutually orthogonal. These assumptions make it possible to approximately reconstruct feature coefficients via inner products, with
$$a_i \approx  \langle x, v_i \rangle$$
%$$a_i \approx f_i(x) := \langle x, v_i \rangle$$

This mechanism is good way to represent unordered \textit{sets} of features. But how might we compose these vectors to express structures more complicated than sets? One natural way to combine two feature vectors $x$ and $y$ is simply to add them up. The sum $x + y$ represents something like the union of the features each individual vector encodes. From a geometric point of view, the sum is a vector that is similar to both $x$ and $y$. This has obvious utility for representations: for instance, one could describe a scene with a duck and a capybara by adding a ``duck vector'' to a ``capybara vector''.

There's an essential limitation to summation, however, which is that there is no sense of order or other structure. A scene where a duck is perched on a capybara and one where the capybara is perched on the duck would have the same representation. Analyzing this issue leads to deep waters. Indeed, Fodor and Pylyshyn famously argued that neural nets are intrinsically limited in their ability to express more complex, directed relations \cite{fodor1988connectionism}. There have been many, many responses to that argument. One particular type of counterargument is to suggest a mechanism for composing or binding elements in a way to preserves relationships.

Most of these relational composition mechanisms fall into three broad classes, depending on whether they're based on binding vectors with matrices, tensors, or references across tokens. This division doesn't capture the full range of possibilities, but it's enough for our purposes. We now discuss some of these mechanisms, and their potential implications for finding and using features.

\section{Additive matrix binding}

One general class of composition mechanisms is based on matrix transformations. 
Suppose $x, y \in \mathbb{R}^n$ are two  vectors. How might we represent the ordered pair $(x, y)$ using a single vector $r \in \mathbb{R}^n$? As discussed, we can't just use the sum $x + y$ since this would destroy any sense of order: the pairs $(x, y)$ and $(y, x)$ would have the same representation.

One method that does preserve order is to fix two distinct $n \times n$ matrices $A$ and $B$ and define the representation of $(x, y)$ by
$$r = Ax + By$$
It's clear that $(x, y)$ and $(y, x)$ will generically have different representations. Moreover, with reasonable assumptions we can generally recover the individual features of both $x$ and $y$ from this representation. If $A$ and $B$ are random orthogonal matrices, and we've assumed that the feature vectors $\{ v_i \}$ are mutually close to orthogonal, it follows that the feature vectors $\{Av_1, \ldots, Av_m, Bv_1, \ldots, Bv_m \}$ will also be close to mutually orthogonal. 
The data for $x$ and $y$ can then be ``read back'' selectively, for example by reconstruction functions of the form
$$f_{A,i}(r) = \langle r, Av_i \rangle$$
$$f_{B,i}(r) = \langle r, Bv_i \rangle$$
Of course, there will be an increase in reconstruction error, since we are working with twice as many features. This issue may become significant if we move beyond pairs of vectors to encoding much longer lists. Thus the addition model of composition might put constraints on $m$, the number of individual feature vectors. On the positive side, the assumptions on $A, B$, and the $\{v_i\}$ can be relaxed in various ways, and useful types of reconstruction may still be possible. For random matrices and most choices of $v_i$, these reconstruction functions will also work. If $A$ and $B$ are random but low-rank, full reconstruction won't be possible for all $v_i$, however it will be possible when $\Vert Av_i \rVert| \approx \Vert v_i \rVert = 1$.

This technique has been proposed in many different contexts. It is called an ``additive model'' in \cite{mitchell2008vector, mitchell2010composition}, a ``generic additive model'' in \cite{zanzotto2010estimating}, and a closely related method is called ``matrix binding of additive terms'' in \cite{gallant2013representing}. It is simple and flexible, naturally generalizing to represent ordered lists of size larger than two. It also has obvious parallels with the output of an attention layer of a transformer.

\subsection{Matrix binding as writing to ``slots'' in superposition}

There's an alternative perspective which relates directly to transformers. Consider an additive matrix binding model that combines a vector $x$ with a list of vectors $(y_1, \ldots, y_k)$ via the representation:
$$r = x + A_1y_1 + A_2y_2 + \ldots + A_ky_k$$
One way to view this equation is that each matrix $A_i$ defines a distinct ``slot'' where data can be written for later retrieval. In other words, the matrix binding formula augments the information in $x$ with the information in the $\{y_i\}$, such that data in one slot can be distinguished from data in another.

By construction, attention layers perform something not unlike matrix binding\footnote{This analogy is suggestive, but not exact. Three important differences from classic matrix binding are nonlinearities (activation functions and layer norm), the low rank of individual attention output matrices, and the lack of guarantee that different output matrices are nearly orthogonal or far from the identity.}. 
There's a potentially fruitful connection between this setup and the proposal that the transformer residual stream is a communication channel \cite{elhage2021mathematical, merullo2024talking} between different network layers, which can augment it with new data or read information from previous layers. It can be tempting to imagine that the residual stream $S$ is a sum of vector spaces $S = E_1 \oplus E_2 \oplus \, \ldots, \oplus E_k$ where each $E_i$ is a separate information ``channel.'' 
However, the additive model is a reminder that channels themselves might be in superposition, with no clear geometric separation between them\footnote{Arguably, positional information in a transformer is encoded in just this way.}. 

Finally, at a theoretical level, it's worth noting that the formula $Ax + By$ uses a form of superposition to create composition. As pointed out by Olah \cite{olah2024distributed}, in reviewing ideas of Thorpe \cite{thorpe1989local}, superposition and composition are typically in tension. In effect, the additive model is composing two dense codes by superposition. Of course, the slot-based model represents a slightly different type of ``compositionality.'' But in Thorpe's terms one may view this as akin to a dense code with compositional characteristics, something in between a semi-distributed and highly distributed code\footnote{
One recent piece of empirical work is potentially related: \cite{engels2024not} describe ``feature manifolds,'' placed in superposition using linear transformations that satisfy a certain mutual orthogonality condition. This suggests that neural networks are capable of learning something like additive binding mechanisms.}.

\subsection{Tree representations using matrices}

Additive binding can be used to create data structures that are more complex than a collection of slots. For example, here's one possible way to represent a binary tree. Fix two random $n \times n$ matrices, $M_1$ and $M_2$. Consider a parent node $p$ with two children represented by vectors $c_1$ and $c_2$. Then define the representation
$$r_p = M_1c_1 + M_2c_2$$
Applying this recursively, we can find a vector representing each node of the tree; and because generically $M_1M_2 \neq M_2M_1$, distinct nodes will be associated with a distinct vectors. While we have not found this mechanism explicitly mentioned in the literature, it can be viewed as a trivial linear version of a Tree-RNN, a network that creates a vector representation of a tree by operating recursively on its nodes \cite{socher2013parsing, socher2014grounded, bowman2015recursive}.

Tree-RNNs enjoyed a surge of interest in the years before transformers were introduced: using Tree-RNNs to process parse trees seemed to improve performance in multiple systems. Interestingly, for small fixed-depth trees there's a natural way for transformers to implement something like a Tree-RNN, since attention heads can focus on---and thus can read information from---syntactically dependent tokens \cite{phang2019attention, ravishankar2021attention}. It might be worthwhile to look for evidence of this type of representation in real-world networks.

\subsection{Feature multiplicity in additive models}

If neural networks use one of these matrix-based mechanisms, how might that affect the search for features? In an idealized case, there's actually a clear answer: it will create a series of duplicate features, which we call \textbf{echo features}. 

To see this, consider a set of random unit feature vectors $\{v_1, \ldots, v_m\}$, and a model in which a representation vector $x$ is a random sparse linear combination of these vectors. To be concrete, take $x = \sum_1^m a_iv_i$ where the $a_i$ are each independent Bernoulli random variables with $p = 1 / m$. Given a large sample of vectors from this distribution, a capable dictionary learning algorithm, such as a sparse autoencoder, will recover the underlying feature vectors $\{v_i\}$.

Now consider a simple version of the additive model described above. Suppose we apply dictionary learning techniques to representations of the form $z = x + Ay$, where $x$ and $y$ are independently drawn from the same sparse random distribution. If $A$ is random, the vectors $\{ Av_i \}$ will probably be nearly orthogonal to the $\{ v_i \}$. That means we can view $z = x + Ay$ as a sparse sum of $2m$ random vectors, where coefficients are independent Bernoulli with $p = 1/m$. Therefore, applying a dictionary learning algorithm to a large sample of $z$ is likely to identify $2m$ feature vectors 
$$ \{ v_1, \ldots, v_m, Av_1, \ldots, Av_m \} $$
The issue, of course, is that each pair of vectors $(v_i, Av_i)$ actually corresponds to the same underlying feature---the only difference being its position in the ordered pair $(x, y)$. We refer to multiple distinct vectors for the same concept as ``echo'' vectors since they echo each other's representation. Obviously, the problem is not just restricted to pairs: a representation of a list of three vectors would yield triple multiplicity. The tree representation described above could yield even more.

It's an open question whether this idealized setting reflects the behavior or real-world networks.
However, the general structure of transformer attention suggests that something like this phenomenon could easily occur. In fact, one recent report describes how an SAE appears to uncover a large set of ``induction features'' corresponding to predicted next tokens \cite{kissane2024}. A closely related issue is also documented in a real-world setting in \cite{makelov2023subspace}.

If this type of feature multiplicity does turn out to be a common phenomenon, it could make interpretability work more difficult. At the very least, it would greatly expand the catalog of seemingly independent features found by an SAE. Understanding the structure of this catalog would require knowledge of the underlying matrices that lead to echo vectors. Conversely, knowing these matrices would greatly help us organize and analyze SAE-derived features.

\subsubsection{Feature multiplicity and predict/control discrepancies}

Feature multiplicity might cause anomalies when using features to steer a network's behavior during inference. Indeed, a number of recent reports have described systems where the optimal vectors for predicting behavior and steering it turn out to be different \cite{zou2023representation, marks2023geometry, li2024inference,  chen2024designing}. We call this \textbf{predict/control discrepancy}. On its face, this discrepancy is not necessarily shocking---a linear probe can easily pick up on spurious or redundant correlations, such as non-causal aspects of the input data. In \cite{bricken2023towards}, for example, the authors perform experiments to check that features have expected causal effects on the networks output. On the other hand, it is conceivable that feature multiplicity may make predict/control discrepancies especially likely\footnote{See also \cite{marks2024whats} for a related discussion; we return to ideas from that essay in a subsequent section.}.

Here's a hypothetical example that serves as a plausibility argument for this idea. Consider a transformer layer that does nothing but augment the residual stream for a given token with information about the previous token. Imagine it does so with additive matrix binding. In particular, if $t_i$ and $t_{i-1}$ are sequential tokens at layer $N-1$, then after layer $N$ the residual stream at token $i$, which we denote by $t'_i$ becomes
$$t'_i = t_i + At_{i-1}$$
where $A$ is an orthogonal matrix far from the identity. Now imagine that representations in the residual stream for layer $N-1$ involve a feature vector $v_{tox}$ for toxic content. It follows that, for different tokens, the residual stream in layer $N$ could have multiple distinct vectors related to toxicity: $v_{tox}$, $Av_{tox}$, or $v_{tox} + Av_{tox}$. A linear probe $P_{tox}$ trained to recognize toxic content, based on multiple tokens in the residual stream at layer $N$, might look something like
$$P_{tox}(x) = \langle x, v_{tox} + Av_{tox} \rangle$$
On the other hand, if we trained on just one token (perhaps the last token in a prompt, a non-toxic function word, in order to predict subsequent behavior) we could easily produce a probe such as:
$$Q_{tox}(x) = \langle x, Av_{tox} \rangle$$
Both of these probes might perform well, but this already illustrates a problem: different tokens might easily produce different probes for unimportant contextual reasons.

Now, suppose we wish to intervene during network inference---for instance, we want to change the residual stream to cause the network to produce less toxic text.
Simply inspecting the second probe, we might hope to intervene to ``remove toxicity'' by intervening during inference \cite{li2024inference} to change each token $t$ to have the new value 
$$\tilde{t} = t - Av_{tox}$$
It's entirely possible, however, that the values of inner products $\langle x, v_{tox} \rangle$ and $\langle x, Av_{tox} \rangle$ will play different roles in computing the toxicity of the next word. It's certainly conceivable that the toxicity of the current word plays a greater role than information about the previous word. Quite likely, the most effective intervention will look more like the following equation, where $c_1 \neq c_2$:
$$ \tilde{t} = t - c_1v_{tox} - c_2Av_{tox}$$
In other words, the right vector for \textit{controlling} the network might differ from the most effective probe vector for \textit{predicting} its behavior.

This phenomenon would make it harder to exploit features derived from dictionary learning. For example, if there are many multiple features related to toxicity, one couldn't automatically extract a steering vector from the dictionary, but would need to do experiments to find the right linear combination of feature vectors.
Of course, the analysis outlined here is still speculative. On the other hand, we believe this is a plausible picture, especially given transformer architecture. A cynic might even say attention layers are machines for producing spurious correlations.

\section{Multi-token mechanisms}

So far, we've discussed binding mechanisms that operate wthin a single vector space. However, since the earliest days of the field, there has been speculation that that some notion of sequence---that is, representations across multiple input observations---is necessary for representing complex relationships \cite{rosenblatt1961principles}. In this section, with the transformer architecture in mind, we discuss a class of mechanisms that relate specifically to sequence models.

Mathematically, we consider a sequence of ``token vectors'' $$t_1, t_2, \ldots, t_k \in \mathbb{R}^n$$
where each $t_i$ contains feature information and, potentially, information about its relation to other tokens. For instance, each $t_i$ might represent the residual stream for the $i$-th element of a sequence at a particular layer in a transformer. To extract the relational data, it might not be enough to look at a single token $t_i$ but at pairs of tokens $(t_i, t_j)$. This perspective opens up a different set of possibilities. It's also a very plausible type of representation. For example, when working with word embeddings, a variety of relationships between words can famously be read by taking vector differences~\cite{mikolov2013distributed}.

A generalization of using vector differences is ``linear relational embedding''~\cite{paccanaro2001learning}. The idea here is that a  relation between token vectors $t_i$ and $t_j$ is present when
$$t_i \approx At_j + b$$
for a matrix $A$ and vector $b$ which depend only on the relation. Recent work suggests that something like this representation may be found in the way that language-model transformers represent relationships between entities~\cite{hernandez2023linearity}.

\subsection{Syntactic relations and tree embeddings}

One striking finding related to the BERT network \cite{devlin2018bert} is that parse trees are represented geometrically by relative positions of token vectors. In particular, the ``syntactic distance'' between two words---as measured in a dependency grammar parse tree---can be recovered from their corresponding token vectors in a middle layer of BERT~\cite{hewitt2019structural, manning2020emergent}. To be precise,  if two words $w_i, w_j$ in the same sentence are represented by embedding vectors $t_i$ and $t_j$, and if $d_s(w_i, w_j)$ is the tree distance in the syntactic parse tree between  $w_i$ and $w_2$, then 
$$\lVert Mt_i - Mt_j \rVert^2 \approx d_s(w_i, w_j)$$
where $M$ is a constant linear transformation, depending only on the network. The fact that tree distance corresponds to the square of the Euclidean distance may seem surprising but, as described in \cite{reif2019visualizing}, it's actually natural in the context of mapping a tree metric to a Euclidean metric. As described in \cite{chi2020finding}, one possible explanation of this representation is that the differences between token embeddings represent syntactic dependencies: that is, if $t_i$ and $t_j$ have a specific syntactic relation (such as an adjective modifying a noun) then the vector difference $Mt_i - Mt_j$ encodes that relation.

This area seems like a promising direction for follow-up work. For one thing, tree structures might naturally occur in many other situations. More generally, the idea that differences between tokens encode specific, contextual relationships seems powerful. A natural question is whether applying dictionary learning to residual stream differences, rather than the residual stream itself, might yield a set of interpretable ``relation features.'' 

\subsection{Reference mechanisms: pointers and identifiers}

Software engineers have invented many ways to tie different data structures together. Is it possible that neural networks use the same techniques? Some recent investigations hint that they might.

One fundamental component of many software data structures is a \textbf{pointer}: a reference to a location in memory. The analog of a pointer, for a transformer network, might be a positional embedding that defines a reference to a specific token in a sequence. One might imagine using matrix binding to augment a token with this positional information rather than semantic data. For instance, if $p_j$ represents the positional embedding for position $j$, and $A_r$ is a binding matrix, one might represent a relationship between tokens $t_i$ and $t_j$ as:
$$r(t_i, t_j) = t_i + A_rp_j$$
In fact, the study described in \cite{prakash2024fine} uncovers a circuit that seems to use something like this kind of ``pointer'' mechanism to relate an entity to information about its state elsewhere in a sequence. The authors describe specific attention heads that seem to move and read relevant positional encodings in order to connect tokens in this way.

A second fundamental way to connect data comes from the world of databases: using a shared \textbf{identifier}, or ID, to link two pieces of information. Some transformers may actually use a geometric form of this mechanism \cite{feng2023language, feng2024monitoring}. The idea is that there may be a special subspace of the residual stream in which vector similarity encodes binding information. To be precise, they find a certain low-rank matrix $A_\mathrm{id}$, with the property that for a token vector $t$, the value $A_\mathrm{id}t$ acts like an ID. That is, two token representations $t_1$ and $t_2$ are considered to be bound to each other if $$A_\mathrm{id}t_i \approx A_\mathrm{id}t_j$$
One interesting question is how and whether this idea relates to the syntax tree representations described in \cite{manning2020emergent}. For example, after applying the proper structural probe, one could imagine projecting orthogonally to an adjective-modifies-noun direction. This could produce something like an ID vector linking adjectives with the nouns they modify. 

A geometric ID mechanism may pose challenges for a pure feature-vector-based analysis. In particular, there's no reason to assume that ID information clusters in any useful way. The only structure that \cite{feng2023language} note in the subspace $S$ is metric: nearby vectors are more likely to match as markers. Thus for any given sequence, ID vectors conceivably could be effectively random---the only constraint being that distinct ID vectors should be far from each other. Thus dictionary learning or classic linear probing may not encounter a useful signal.

\section{Vector symbolic architecture}

Finally, we discuss a set of historical ideas for vector binding, collectively known as ``vector symbolic architecture,'' or VSA \cite{smolensky1990tensor, plate1994distributed, plate1997common, kanerva2009hyperdimensional, jones2007representing, schlegel2022comparison}. A comprehensive survey and analysis of the zoo of techniques in this area is far beyond the scope of this paper. Our goal is simply to provide a gentle introduction to techniques that can potentially seem arcane, unmotivated, and intimidating. To do so, we focus on two foundational constructions, one in $\mathbb{R}^n$ and one in $\{0, 1\}^n$, which underlie much of the work in this field.  Both mechanisms have the potential to lead to feature ``dark matter.'' That is, they might represent information in a way that is out of reach of dictionary learning or probing methods.

\subsection{Tensor constructions}

As described in \cite{schlegel2022comparison}, a technique introduced in \cite{smolensky1990tensor} forms the basis of a large set of different VSA mechanisms. It is often referred to as a ``tensor'' method, but we'll describe it with the more down-to-earth notation of outer products.

As before, suppose we wish to compose, or bind, two vectors $x$ and $y$. One way is to use an outer product to define a representation:
$$r = xy^T$$
This representation distinguishes order, since generally $xy^T \neq yx^T)$. It respects vector addition, since it is linear in each of $x$ and $y$ respectively. Moreover, one can recover $x$ and $y$, at least up to scalar multiples, with easy vector algebra. For example, if you know $y$, you can recover a scalar multiple of $x$ via the product $$ry = xy^Ty = \lVert y \rVert^2 x$$

The alert reader will notice, however, that this outer product representation is a cheat: it lives in $\mathbb{R}^{n \times n}$ rather than $\mathbb{R}^n$. One may view much of the VSA literature as a bag of tricks to get around this inconvenient fact. The most common trick is to create an $n$-dimensional projection of the outer product, which can still be used for approximate reconstruction. One method, for instance, is projecting the outer product matrix via circular convolution \cite{plate1997common}.

\subsection{Binary vectors}

A second category of VSA techniques relies on binary-valued vectors \cite{kanerva1997fully, kanerva2009hyperdimensional}. We describe one simple proposal of this type. To bind two binary vectors $x$ and $y$, we fix a permutation $P$ that rearranges the entries of a vector\footnote{We need the permutation $P$ because mod-2 addition is commutative, and otherwise we couldn't distinguish the pairs $(x, y)$ and $(y, x)$. The situation is similar to an additive matrix binding formula $x + Ay$. Permuting one argument also allows us to bind a vector to itself without leading to a zero representation.}, and then compute
$$r = x \oplus P(y)$$
Here $\oplus$ denotes modulo-2 addition. Like the outer product, this distinguishes between the ordered pairs $(x, y)$ and $(y, x)$. It respects the binary OR function, which can be used to take unions of two sets of features. It also allows for easy reconstruction of one vector in the pair if you know the other:

$$ x = r \oplus P(y)$$
$$ y = P^{-1}(r \oplus  x)$$

This is a conceptually elegant method, although it requires extremely large vectors for effective storage and decoding.

\subsection{Is VSA a plausible mechanism for real neural nets?}

Unlike additive binding models, the techniques proposed for tensor-based VSA don't obviously map to deep learning architectures. Investigations of whether real-world networks use VSA have yielded mixed conclusions. For example, a study of whether CLIP might use these mechanisms produced largely negative results \cite{lewis2022does}, while there are some indications RNNs may reproduce tensor-related structures \cite{mccoy2018rnns}. 

Techniques based on binary vectors might seem even further from modern language model architectures. 
Surprisingly, some recent work suggests that for sparse vectors in superposition, not only can neural nets compute XORs of such features in theory \cite{vaintrob2024toward}, they seem to do so in practice \cite{marks2024whats}. Traditional binary VSA is based on dense vectors, of course, but this seems like an interesting avenue for exploration. One might imagine combining ideas from the tensor and binary VSA schemes---perhaps based on outer products specifically of zero-one vectors, either over $\mathbb{R}$ or $\mathbb{Z}_2$.

\subsection{Implications for the search for features}

As with matrix addition binding, we speculate that VSA mechanisms may lead to feature multiplicity.  For example, outer product binding could plausibly produce a combinatorial proliferation of potential features\footnote{The essay \cite{marks2024whats} has a discussion of parallel concerns in the case of binary codes.}. To see the issue, consider a setting where we have $m$ feature vectors, $\{v_1, \ldots, v_m \}$, and we want to compose representation vectors $x$ and $y$:
$$x = \sum_{i = 1}^m a_i v_i$$
$$y = \sum_{i = 1}^m b_i v_i$$
where the sums are assumed to be sparse, and $\{a_i\}$ and $\{b_i\}$ independently chosen. Suppose we have a composition method of the form
$$ r(x, y) = \pi(xy^T)$$
where $\pi: \mathbb{R}^{n \times n} \rightarrow \mathbb{R}^n$ is a projection operator.
By linearity, we have
$$r(x, y) = \sum_{i, j} a_ib_j\pi(v_iv_j^T)$$
We have now moved from using $m$ feature vectors $\{v_i \}$ to a situation with $m^2$ potential feature vectors $\{ \pi(v_iv_j^T) \}$.

This is a problem. To begin with, even if we could use dictionary learning to find all these features, there may be a truly huge number of them. Even if not all combinations are likely to occur, the potential number of new combinatons is vast. 
For a person interpreting a network, having fine-grained versions of the same feature is confusing. For example, one might see many different vectors reflecting the same concept in different overall contexts. Something like this actually appears repeatedly in a language model analyzed in \cite{bricken2023towards}. For example, they don't find a single pure ``English preposition'' feature, but they do identify a feature corresponding to ``prepositions in scientific/statistical contexts''\footnote{Feature A/1/3533 in the study's notation} and a separate feature that fires on ``prepositions in contexts discussing poetry/poets''\footnote{Feature A/1/3977 in the study's notation}. This is just one of many examples from the feature catalog of \cite{bricken2023towards}. As the authors of that study discuss, this seems to be part of a much larger story related to a non-isotropic distribution of features---perhaps an analysis of compositional mechanisms could shed light on this phenomenon.

A second problem with the proliferation of fine-grained features is that we might need far more data to discover each one. For example, features for ``inspiring'' and ''architecture'' might be just common enough to find, but the combination of ``inspiring architecture'' might, sadly, be too rare to appear in the output of a dictionary learning algorithm. This may lead to ``dark matter''---feature representation that are out of reach unless one tests with vast amounts of data.

From an optimistic viewpoint, this phenomenon can also be seen as a cause for hope. As discussed earlier, many features in the catalog of \cite{bricken2023towards} seem to have combinatorial interpretations. If we had a principled way of ``factoring'' these features, extracting an outer-product structure automatically, that could be a huge help in identifying primitive features. It could also help simplify and organize feature catalogs.

\section{Conclusion and future work}

We've described a series of proposals---some historical, some very recent---for how neural nets might represent relationships between features. Along the way, we've given conceptual arguments that these mechanisms might present problems when finding and using linear feature representations. One challenge is that of ``dark matter'': feature representations that are difficult to find using standard methods. Another potential problem is feature multiplicity, or the presence of multiple ``echo vectors'' that correspond to the same feature. An issue of special concern, related to multiplicity, is whether feature vectors discovered by probing or dictionary-learning methods will be as useful for interventions as they are for predicting the state of the network. 

The arguments here are conjectural and often involve idealized settings. A great deal of work would be necessary to resolve the questions we've raised---calling for that work is the point of this note. Here are some directions that might shed light on the key issues:

\begin{itemize}

  \item \textbf{Toy models that learn composition}. One way to gather data would be to experiment with minimal models. It would be useful to find toy models of learned composition. If we give an autoencoder a task that requires learning relational composition, what mechanisms are found via gradient descent? Some work in this area dates back decades \cite{pollack1988implications, blank2014exploring} and it would be interesting to analyze these same systems with modern methods, and to extend this line of research further.
  \item \textbf{Apply feature extraction methods to synthetic composition mechanisms}. We provided a conceptual outline of how feature discovery techniques might run into trouble on compositional representations. A natural next step would be to test these arguments empirically, by creating synthetic versions of VSA, matrix binding, and other mechanisms, and applying dictionary learning to the results. 
  \item \textbf{Apply dictionary learning to token differences}. Syntactic relations appear to be encoded in differences between token vectors. Are other relations encoded this way as well? It could be worth applying dictionary learning techniques to token differences to find additional ``relational features.'' 
  \item \textbf{Investigate marker mechanisms}. The ``ID vector'' mechanism suggested by \cite{feng2023language} is extremely interesting. It also presents a clear problem for feature-based interpretability methods. It would be helpful to understand just how widely the mechanism might apply. How does it relate to the syntax representations of~\cite{manning2020emergent}? Experimentation with small synthetic models might advance our understanding of the basic process. It would also be helpful to find lightweight ways to identify ID vector subspaces.
  \item \textbf{Look for feature multiplicity}. If feature multiplicity is a real problem, it may not be hard to identify. One sign would be the discovery of redundant or contextually dependent feature vectors. The features found in \cite{bricken2023towards, templeton2024scaling, kissane2024} sometimes exhibit this type of behavior, for instance. If one could pair up a sufficiently large number of redundant features via a single linear transformation, that would provide strong evidence for the multiplicity hypothesis, and also help organize feature catalogs. 
  \item \textbf{Understand predict/control discrepancy}. Feature multiplicity might cause a problem that we have termed ``predict/control discrepancy.'' That is, an effective vector for predicting a particular network behavior might be different from the best ``steering vector'' for inducing that same behavior. Given the obvious practical implications, and that there are multiple examples of this phenomenon in the literature, it seems important to investigate further.
  \item \textbf{Measure real models for relational composition}. One alternative to explicitly identifying relational representations is to look for behavioral evidence that they exist. That can at least alert us to ``dark matter'' that isn't observed by feature identification methods. A review of this type of work is beyond the scope of this paper, but several studies suggest promising directions \cite{andreas2019measuring, lovering2022unit, akyurek2023lexsym}.
  \item \textbf{Investigate VSA-based binding in real networks.} The findings in \cite{vaintrob2024toward, marks2024whats} suggest some natural mechanisms by which binary VSA codes might be implemented in realistic networks. It may be worth systematically working out the simplest implementations of these mechanisms, and investigating whether real networks make use of them.

\end{itemize}

To sum up, there are major practical and theoretical questions around how the linear representation hypothesis might interact with mechanisms for relational composition. We strongly advocate for continued research in this area. Identifying mechanisms for relational composition would be a step forward for useful interpretability work. Even if it turns out that neural nets show no evidence of relational composition, that would be an important theoretical result.

\section{Acknowledgments}

We thank Kenneth Li, Oam Patel, Nikola Jurkovic, Chris Olah, Yonatan Belinkov, Asma Ghandeharioun, Ann Yuan, and Lucas Dixon for helpful comments on this manuscript. We're indebted to the anonymous reviewers for their careful reading and excellent suggestions. FV was supported by a fellowship from the Radcliffe
Institute for Advanced Study at Harvard University. Additional support came from
Effective Ventures Foundation, Effektiv Spenden Schweiz, and the Open Philanthropy Project.

\bibliography{references}
\bibliographystyle{icml2024}

%%%%%%%%%%%%%%%%%%%%%%%%%%%%%%%%%%%%%%%%%%%%%%%%%%%%%%%%%%%%%%%%%%%%%%%%%%%%%%%
%%%%%%%%%%%%%%%%%%%%%%%%%%%%%%%%%%%%%%%%%%%%%%%%%%%%%%%%%%%%%%%%%%%%%%%%%%%%%%%
% APPENDIX
%%%%%%%%%%%%%%%%%%%%%%%%%%%%%%%%%%%%%%%%%%%%%%%%%%%%%%%%%%%%%%%%%%%%%%%%%%%%%%%
%%%%%%%%%%%%%%%%%%%%%%%%%%%%%%%%%%%%%%%%%%%%%%%%%%%%%%%%%%%%%%%%%%%%%%%%%%%%%%%
\newpage
\appendix
%\onecolumn

\section{Appendix: Features}

In this paper we use the word ``feature'' formally, as a kind of mathematical abstraction, since that's all that's needed for our arguments. However, it's worth talking about some of the intuition behind the word. To begin with, there's not a consensus definition in the literature, which is too large to survey in an appendix. See \cite{elhage2022toy}, for example, which discusses three approaches: features as arbitrary functions of the input, features as interpretable properties, and---most abstractly---features as what an infinitely large neural net might devote a single neuron to.

The general intuition we've found most helpful, however, is simply that a feature represents a unit of data that is useful for future computations. We also believe it's helpful to divide these representations into three broad categories. Some features relate to the input---that is, they represent a useful property of the data seen by the network. (``Red area on green background'' or ``apple.'') Other features, like data structures in a traditional algorithm, are related to intermediate processing. (``Food and hunger''.) Finally, a third set of features will relate to the result of the network's computations: they will represent properties of the output, rather than input (``Reach arm'' or ``open mouth''). 

The analogy is obviously with biological nervous systems, which can be broken into areas for sensory, motor, and internal processing. Of course, there's no guarantee that a neural network maintains a strict division between these different aspects. It's conceivable that within a given layer, all three types of features may be found. That may be another reason why interpreting features can be hard. Frequently people try to understand features in terms of input data, when those features may really be aspects of hidden computation or output. It may also explain some predict/control discrepancies: linear probes may be effectively trained in a way that emphasizes features of the input, which are only roughly correlated with output features.
%%%%%%%%%%%%%%%%%%%%%%%%%%%%%%%%%%%%%%%%%%%%%%%%%%%%%%%%%%%%%%%%%%%%%%%%%%%%%%%
%%%%%%%%%%%%%%%%%%%%%%%%%%%%%%%%%%%%%%%%%%%%%%%%%%%%%%%%%%%%%%%%%%%%%%%%%%%%%%%

\end{document}